%% file: PaperForReview.tex
\documentclass[10pt,twocolumn,letterpaper]{article}

\usepackage{cvpr}              

\usepackage{graphicx}
\usepackage{amsmath}
\usepackage{amssymb}
\usepackage{booktabs}
\usepackage{algorithm}
\usepackage{algpseudocode}
\usepackage[table]{xcolor}
\usepackage{pifont}

\usepackage[pagebackref,breaklinks,colorlinks]{hyperref}

\usepackage[capitalize]{cleveref}
\crefname{section}{Sec.}{Secs.}
\Crefname{section}{Section}{Sections}
\Crefname{table}{Table}{Tables}
\crefname{table}{Tab.}{Tabs.}

\def\cvprPaperID{*****} 
\def\confName{CVPR}
\def\confYear{2022}

\def \ninstances {$2,398$}
\def \nplots {$388$}

\def \dataset {\texttt{\textbf{\textit{SICKLE}}}}

\newcommand{\cmark}{\ding{51}}%

\begin{document}

\title{High-Resolution Satellite Imagery for Modeling the Impact of Aridification on Crop Production}

\author{Depanshu Sani$^1$\\
{\tt\small depanshus@iiitd.ac.in}
\and
Sandeep Mahato$^2$\\
{\tt\small sandeepmahato@mssrf.res.in}
\and
Parichya Sirohi$^1$\\
{\tt\small parichays@iiitd.ac.in}
\and
Saket Anand$^1$\\
{\tt\small anands@iiitd.ac.in}
\and
Gaurav Arora$^1$\\
{\tt\small gaurav@iiitd.ac.in}
\and
Charu Chandra Devshali$^2$\\
{\tt\small ccdevshali@mssrf.res.in}
\and
Thiagarajan Jayaraman$^2$\\
{\tt\small jayaraman@mssrf.res.in}
\and
Harsh Kumar Agarwal$^1$\\
{\tt\small harsh19423@iiitd.ac.in}
\and
$^1$ Indraprastha Institute of Information Technology, Delhi, India\\
$^2$ M S Swaminathan Research Foundation, Chennai, India\\
}
\maketitle

\input{sections/introduction.tex}
\input{sections/methodology.tex}
\input{sections/analysis.tex}
\input{sections/conclusion.tex}





{\small
\bibliographystyle{ieee_fullname}
\bibliography{egbib}
}

\clearpage
\input{sections/Appendix.tex}

\end{document}

%% file: sections/introduction.tex
\section{Introduction}
\label{sec:intro}

The Cauvery Delta is a major rice cultivation region in Tamil Nadu, India, responsible for the food security and livelihoods of millions of farmers. In a given agricultural year, the farmers from the delta cultivate one or two crops of paddy, depending upon the availability of water. Over the years, conditions have changed in the delta with reports of paddy yield stagnation and uncertainty in water availability for cultivation. An Indian Council for Agricultural Research (ICAR) study in $2013$ reclassified the four districts of Cauvery Delta from dry semi-humid to semi-arid conditions \cite{ICAR_report}. 

In the context of these changes reported for the region and its centrality to rice production, we aim to do a comprehensive analysis of the impact of aridification on paddy yields in the Cauvery Delta by monitoring the changes in cropping pattern and productivity. The type of crop, its sowing \& harvesting dates, evapotranspiration \& yield estimates, etc. can provide insights about the study region. There exist methods that employ single-image or multi-image input \cite{dakir2020crop, singha2016object, zhang2018mapping, ye2018review, dela2021remote, sustainbench, cropharvest, wang2022parcel, palanivel2019approach, clauss2018estimating, sharma2020wheat} which can be used to estimate some of these parameters individually. Even though there are methods that use a sequence of historical images of a study region, they do not take into account the standard practices that are followed in that specific region. They use an arbitrary length of the sequence to generate time-series data. For example, each sequence in PASTIS-R \cite{pastis-r} contains observations taken between $\text{Spetember }2018-\text{November }2019$. Multiple crops have likely been cultivated in this time span and hence the quality of input is degraded $-$ ultimately affecting the performance of downstream tasks. Unfortunately, no work has been done to develop methods that are consistent with the standard agricultural practices followed in a specific region. Moreover, the current literature fails to present an end-to-end framework that can be utilized to predict all (or some) of these parameters. Most of the related work does not make the dataset publicly available and furthermore, there exists no reported dataset that provides multiple cropping parameters annotated for the same set of plots. 

To tackle the aforementioned challenges, we make the following contributions that can benefit the deployment quality and hence promote the applicability of remote sensing and machine learning in agriculture.
\begin{enumerate}
    \item We introduce a first-of-its-kind dataset, \dataset{} (\textbf{S}atellite \textbf{I}magery for \textbf{C}ropping annotated with \textbf{K}ey-parameter \textbf{L}ab\textbf{E}ls) \cite{sickle}, which contains images from multiple satellites (Landsat-$8$, Sentinel-$1$ and Sentinel-$2$) having a variety of sensors (optical, thermal and radar) with annotations of multiple parameters for each individual plot in the regions of Cauvery Delta in Tamil Nadu, India. These annotations are created at $3$ levels of spatial resolution, i.e. $30m$, $10m$ and $3m$, opening the pathways for research in the direction of high-resolution (HR) inferences from low-resolution (LR) data.
    \item We release the crop sowing, transplanting and harvesting dates of the samples, that motivates the employment of the actual growing season of the crops to improve the performance of downstream tasks. We prove this hypothesis by empirically showing the results of employing actual growing season length on the crop yield prediction task.
    \item We present a novel method for time-series data generation which is consistent with the standard practices that farmers follow in a specific region. Hence, ensuring the robustness and correctness of any solution in a real-world deployment.
    \item We highlight the need for an end-to-end framework that predicts multiple key cropping parameters for a region using a sequence of historical images.
\end{enumerate}

%% file: sections/methodology.tex
\section{Methodology}
\label{sec:methodology}

Our methodology is inspired by the multiple tasks that are presented by Yeh et al. \cite{sustainbench} in \textsc{SustainBench}. We build our codebase on their repository, wherein they provide the data loaders for all the required tasks and one can employ any neural network for training purposes. We use the similar methods that were used to benchmark \cite{sustainbench}. Similar to Rose et al. \cite{croptypemapping}, we early fuse the images from different satellites by concatenating them band-wise for each timestep wherever required. 

\subsection{Standardized Time-Series Data Generation}
\textbf{Regional Standards:} Consider a scenario where we want to classify each pixel in a study region as paddy or non-paddy. Let's assume we select any arbitrary length $t_a$ for creating the time-series data and the duration of the growing paddy season is $t_p$. If $t_a \ll t_p$, then the time-series observations will not be able to capture the complete phenological structure and hence will lack the necessary information. On the other hand, if $t_a \gg t_p$, the input observations might contain multiple paddy seasons and may also witness multi-crop cultivation. Therefore, we strongly believe that the duration of the observations to create time-series data depends on the downstream task to be performed. 
We propose that unlike the previous works that build methods on arbitrary sequence lengths \cite{pixel-set, pastis-r, sustainbench, cropharvest, agriculture-vision, radiant} we need some `standards' for driving a solution to become more robust, trustworthy and logically correct instead of being data hungry. In this study, paddy is the primary crop under consideration and we use the information obtained from Tamil Nadu Agricultural University \cite{tnau} as a standard. We then curate a dataset based on this standard for paddy crop cultivation, as explained in the next section.

\textbf{Selection of the sequence length:} In general, the crop phenology dates gathered from the farmers might be different from the standard paddy season duration of that region. For instance, while the standard growing season starts in September, a farmer might sow the seeds in the month of October depending on the water availability. On the other hand, there might be crops, such as coconut and banana, that can span a complete year for its cultivation. Therefore, we propose a novel method for time-series data generation where we consider the duration of the standard season as the sequence length for creating the time-series data instead of the gathered cropping dates or any arbitrary sequence length. This strategy is based on the fact that even when the actual growing season is unknown for a crop, one will always have the duration of its standard growing season. Specifically, for each of the standard seasons, we consider all the satellite images of a particular plot between that duration as a single sample. We select the duration having the maximum number of days in its growing season duration as the sequence length and consider each sequence index as the number of days since the standard sowing day. Hence, each index in the sequence corresponds to an image acquired on a particular day in the growing season. All the days for which no image was available are padded with 0s. We adopted this technique for time-series data generation and have curated \ninstances{} instances using the gathered field data.

\subsection{End-to-End Framework}
In this section, we assume that the trained models for crop cover map generation, phenology date prediction and yield estimation are readily available. The primary focus of this section is the development of an end-to-end framework that is capable of predicting multiple cropping parameters without imposing a technology barrier for the end-users. 

Algorithm \ref{alg:framework} explains the working of the framework, which only takes the path to a directory of time-series images of a region and the standard growing season to be considered as input. This framework poses no technical dependencies on the end-user and hence is easily accessible by anyone, which is the ideal requirement of a real-world solution. The methods used in Algorithm \ref{alg:framework} are self-explanatory and therefore, their explanation is not a part of the main paper.

\begin{algorithm}
\caption{get\_crop\_parameters($P$, $S$)}\label{alg:framework}
\textbf{Input:} $P \gets$ Path to directory having tif images of a region\\
     \hspace*{2.9em} $S \gets$ Standard growing season and year\\
\textbf{Output:} $Y \gets$ Crop parameter maps
\begin{algorithmic}
    \Require $Y[5, H, W] \gets \{0\}$ 
    \State $\mathcal{X} \gets$ tif\_to\_numpy\_sequence($P$)
    \State $x \gets$ get\_cropped\_inputs($\mathcal{X}$)
    \State $crop\_cover \gets$ predict(`crop\_cover', $x$)
    \State $Y[0] \gets$ merge\_cropped\_inputs($crop\_cover$)
    \State $paddy\_pixels \gets$ get\_paddy\_pixel\_indices($Y[0]$)
    \For{$(row, col) \ in \ paddy\_pixels$}
        \State $patch \gets$ get\_patch(mid=$(row, col)$, dim=$(h,w)$, $\mathcal{X}$)
        \State $sow\_dt \gets$ predict(`sowing', $patch$)
        \State $transplant\_dt \gets$ predict(`transplanting', $patch$)
        \State $harvest\_dt \gets$ predict(`harvesting', $patch$)
        \State $x \gets get\_actual\_season(patch, sow\_dt, harvest\_dt)$
        \State $yield \gets$ predict(`yield', $x$)
        \State $Y[1][row,col] \gets$ $sow\_dt$
        \State $Y[2][row,col] \gets$ $transplant\_dt$
        \State $Y[3][row,col] \gets$ $harvest\_dt$
        \State $Y[4][row,col] \gets$ $yield\_estimate$
    \EndFor \\
    \Return $Y$
\end{algorithmic}
\end{algorithm}

%% file: sections/analysis.tex
\section{Experimental Analysis}
\label{sec:analysis}
During the Mid AI4SG Workshop, we proved the hypothesis that High-Resolution (HR) imagery could improve the performance of the downstream task. A plot-level binary crop classification task (paddy/non-paddy) was chosen for this purpose. Thus, we conduct new experiments based on this assumption. But due to the addition of newer field data after the mid-workshop and the unavailability of more HR images from Planet Labs, we did further experiments using only Landsat-8 and Sentinel-2. Instead of Planet Scope, we used Sentinel-1 in the new experiments. While Sentinel-1 cannot provide high-spatiotemporal resolution data, it is rich in complementary information because of its non-optical sensors, i.e. microwave.

The downstream tasks in this paper are $(i)$ Crop Type Segmentation, $(ii)$ Phenology Date Prediction, and $(iii)$ Crop Yield Prediction. Phenology Date Prediction task is itself composed of $3$ sub-tasks, i.e. sowing, transplanting and harvesting date prediction. We only present an analytical review of the findings in the main paper. The experimental results are presented in the Appendix section. Looking at the results obtained from the conducted experiments, we make the following observations:
\begin{enumerate}
    \item Even though higher spatial resolution data provides more detailed information, spectral resolution plays a vital role in the performance of the tasks. As compared to higher spatial but lower spectral resolution data, a lower spatial resolution along with a higher spectral resolution can improve the performance significantly. For example, the ultra-blue sensor of Landsat-8 (Table \ref{tab:yield-uncommon} of the Appendix), having 30m spatial resolution, performs slightly better than the RGB and NIR band combination of Sentinel-2 (Table \ref{tab:yield-common}), having 10m spatial resolution, for the yield prediction task. In the same table, it is also worth noticing that the results obtained using SWIR-1 and SWIR-2 bands alone perform better than the RGB and NIR band combination (maximum spectral resolution of Planet Scope). Thus we hypothesize that if Planet Scope has these short-wave infrared sensors, they can perform even better. There exist methods to improve the spectral resolution, \cite{rgb2nir, nir-generation} by generating new spectral bands, but these methods need ground truth band values. Recent works on pan-sharpening for improving the spatial resolution are also interesting \cite{pan_comparisons, pan_ICCV, pan_multisensor}; however, these methods are subject to the availability of HR panchromatic bands from the same satellite. An interesting research direction could be to investigate cross-satellite fusion to achieve high spectral resolution.
    
    \item Since certain spectral bands are more sensitive than others for a given downstream task, naively selecting all the spectral bands does not guarantee better performance. For instance, in Table \ref{tab:yield-common} of the Appendix, some of the band combinations perform better than the results obtained by naively selecting all the band combinations. At the same time, certain combinations downgraded the performance. Therefore, there is a need for a learning-based algorithm that can intelligently select the best band combination depending on the primary task under consideration. Unfortunately, to the best of our knowledge, there exists no such method. This is another promising research direction. 
    \item Using multi-sensor data from multiple satellites improves the performance of the downstream tasks, as shown in Table \ref{tab:yield-fusion} of the Appendix section. Even though we achieve better performance using cross-satellite fusion, it must be noted that all these different satellites have varying characteristics, such as pose, orientation, swath width, altitude, sensors and their sensitivity, etc. Geo-registration is another open research problem in this domain. Early fusing the satellite images simply ignores these fundamental differences.  In \cite{MISR-HSI-MSIReview}, the authors report the impact of geo-registration errors on fusion performance. Similarly, solar angles, swath width etc., may also lead to pixel-wise differences between images. Unfortunately, the current literature focuses on data-driven learning methods instead of geometry-aware learning.
    \item Employing the actual growing season for Crop Yield Prediction improves the performance significantly in comparison with the usage of the standard growing season.
\end{enumerate}

%% file: sections/conclusion.tex
\section{Taking it Forward}
\label{sec:conclusion}
We intend to take forward the ongoing research and its deployment to comprehend and capture historical and current vis-a-vis real-time spatio-temporal changes in key crop parameters: crop cover, phenology, yield, and soil moisture demands for the Delta. Immediate tasks includes integration of evapotranspiration estimates and obtaining cropland masks to evaluate land fallowing during the paddy season. Next stages will involve increasing the spatial and temporal scale of the deployment, collection of ground truth data and solving the research gaps discussed in Section \ref{sec:analysis}. We also desire to deploy the model to perform progressive predictions on the crop parameters using real-time images. For a detailed outline of the proposed work forward, one can refer to the Project Narrative Report, July 2022.

%% file: sections/Appendix.tex
\section*{APPENDIX}
\section*{Dataset}
\begin{figure*}[b]
    \centering
    \begin{subfigure}[b]{0.48\linewidth}
        \centering
        \includegraphics[width=\linewidth]{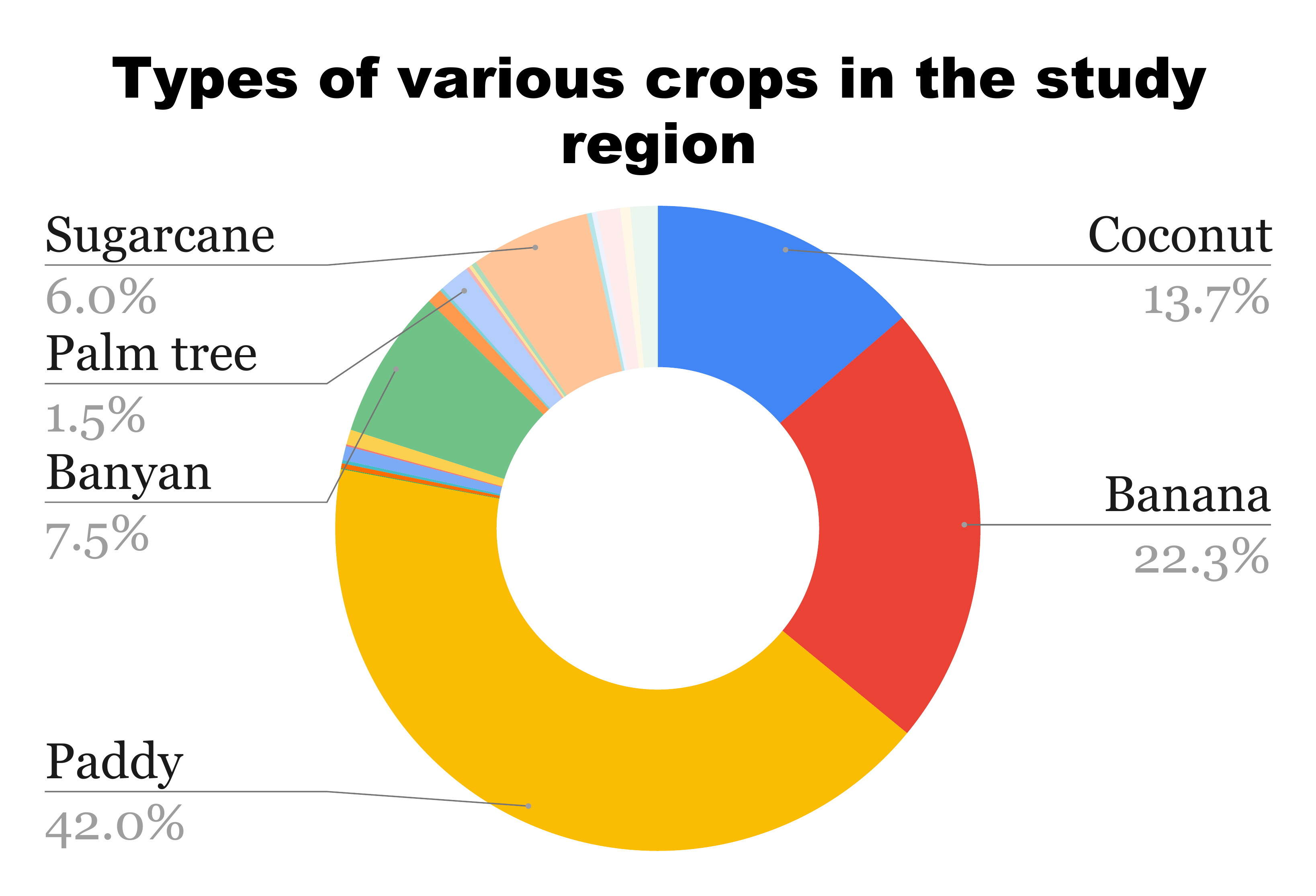}
        \caption{}
        \label{fig:n-crop}
    \end{subfigure}
    \begin{subfigure}[b]{0.48\linewidth}
        \centering
        \includegraphics[width=\linewidth]{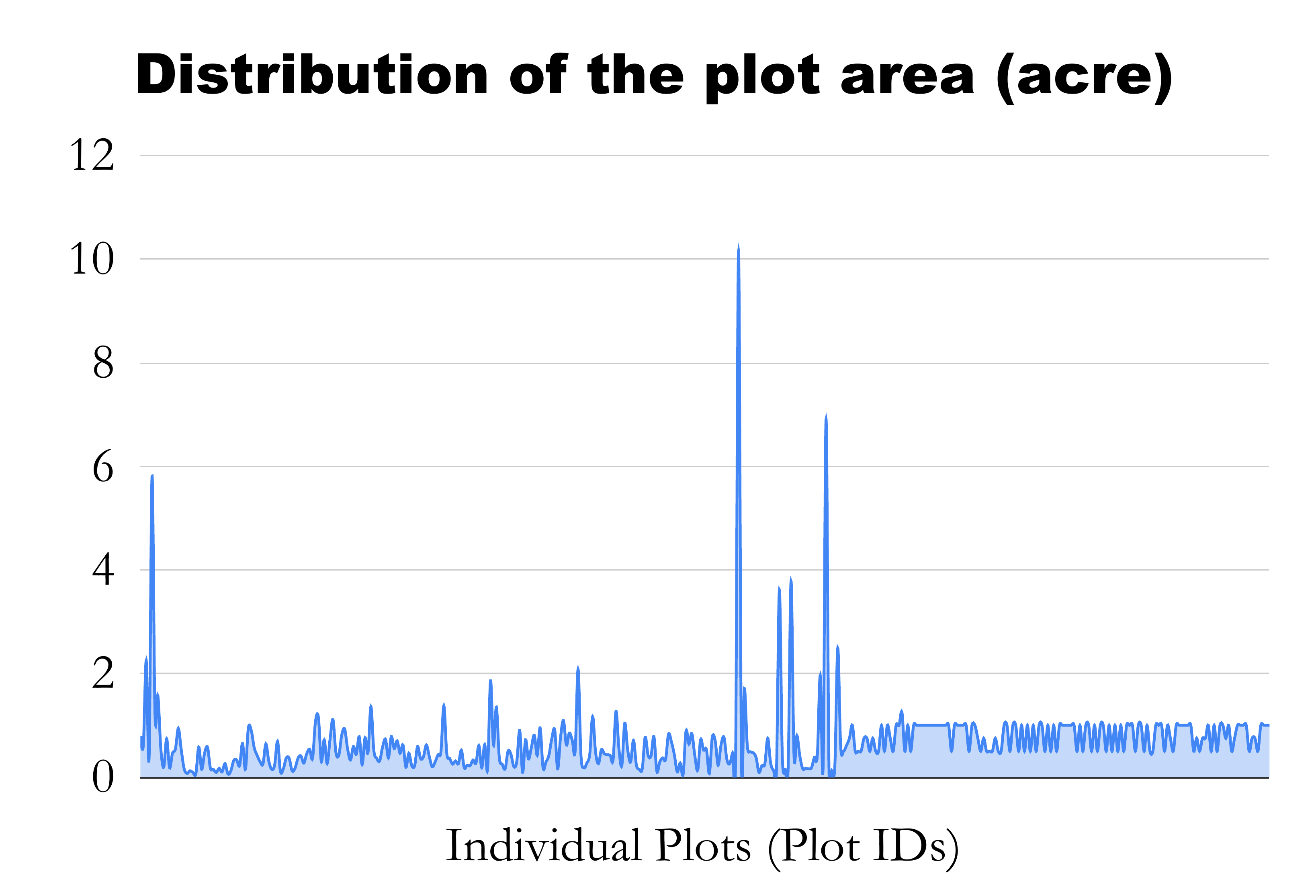}
        \caption{}
        \label{fig:area}
    \end{subfigure}
    \begin{subfigure}[b]{0.48\linewidth}
        \centering
        \includegraphics[width=\linewidth]{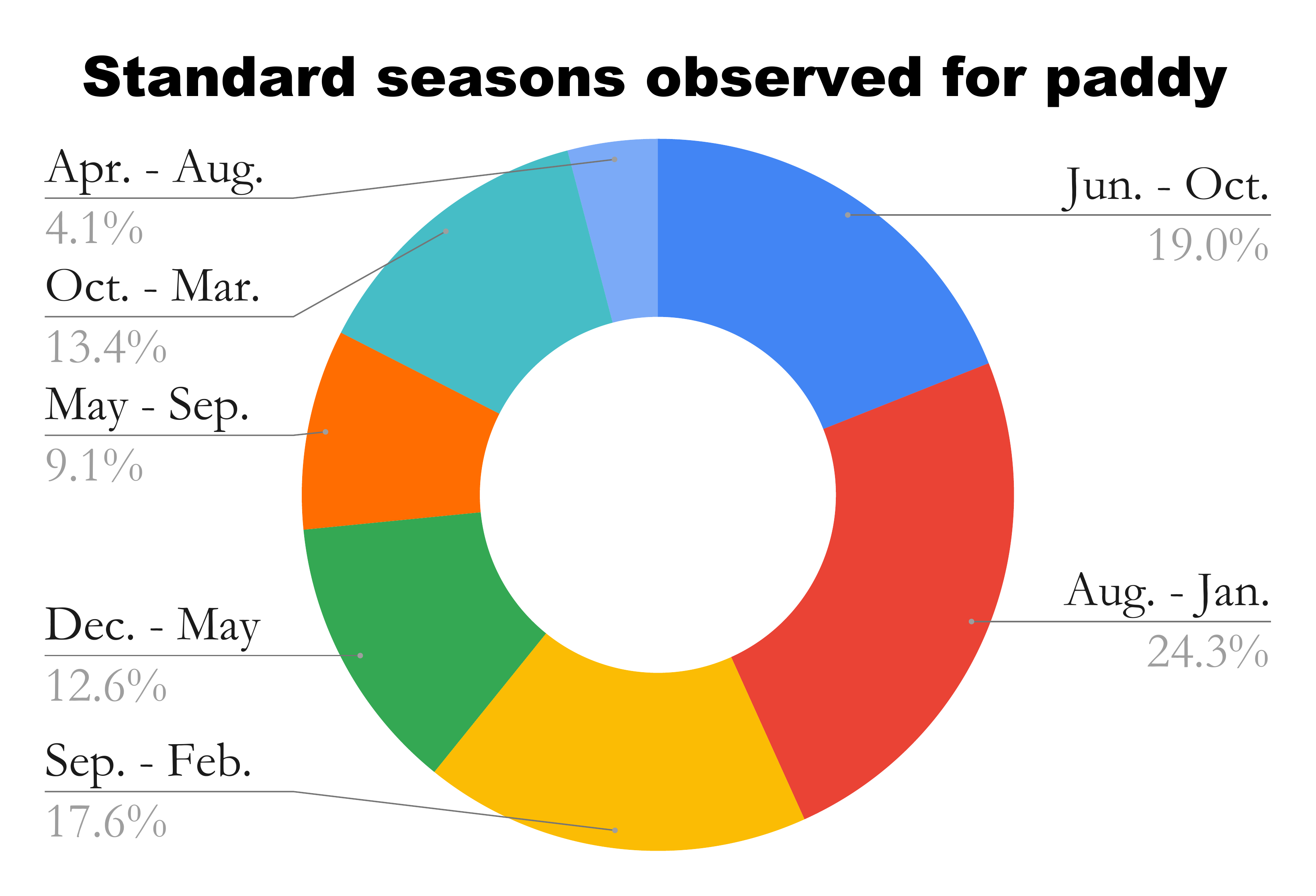}
        \caption{}
        \label{fig:n-seasons}
    \end{subfigure}
    \begin{subfigure}[b]{0.48\linewidth}
        \centering
        \includegraphics[width=\linewidth]{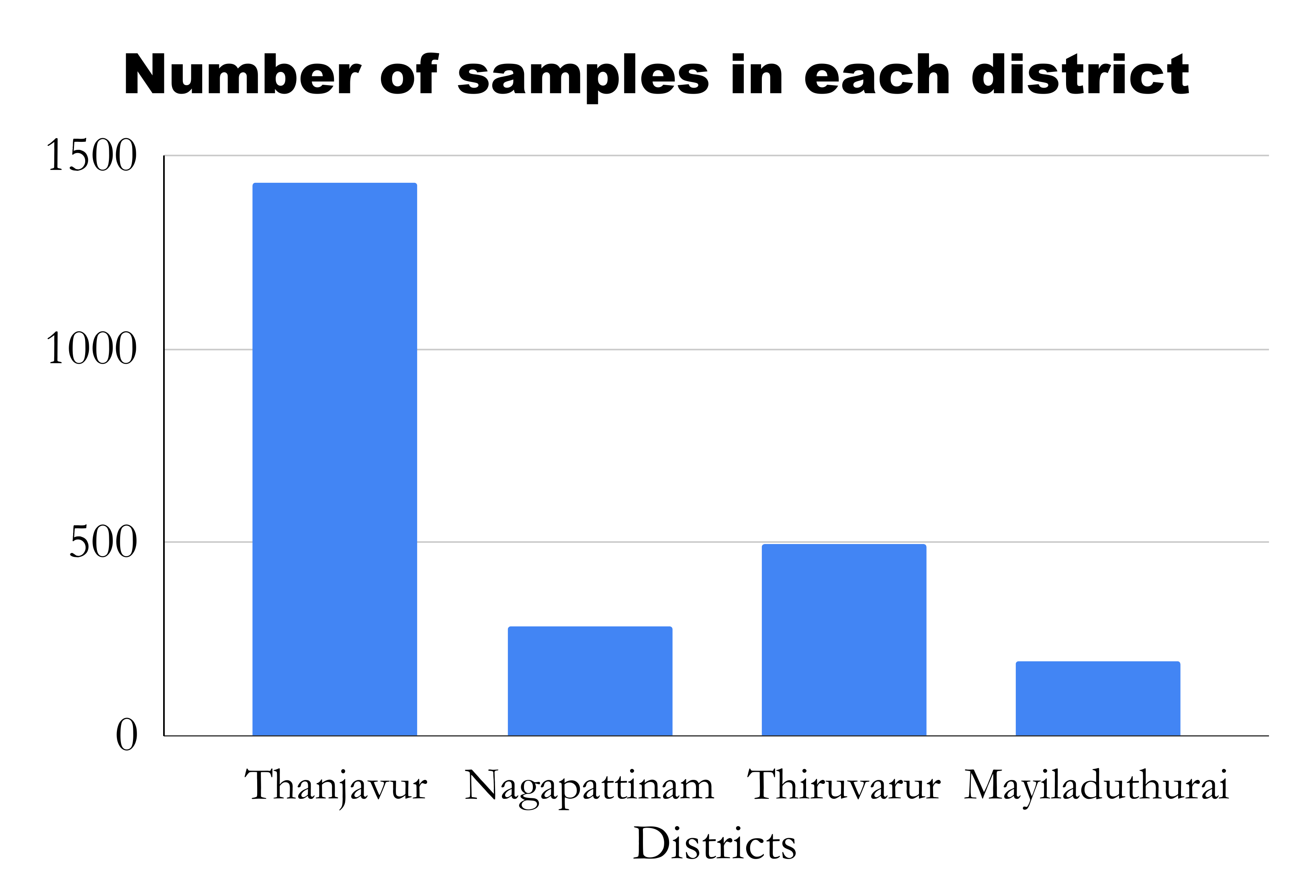}
        \caption{}
        \label{fig:n-samples}
    \end{subfigure}
    \caption{Dataset Statistics. Figure \ref{fig:n-crop} shows the distribution of various types of crops in the dataset that are cultivated in the study region. Figure \ref{fig:area} represents the area (in acre) of each of the \nplots{} individual plots. The X-axis denotes the plot id of each plot and the Y-axis denotes the area (in acres). Figure \ref{fig:n-seasons} shows the distribution of the standard paddy seasons observed in the study region. Figure \ref{fig:n-samples} shows the district-wise distribution of the collected \ninstances{} samples.}
    \label{fig:dataset-analysis}
\end{figure*}

\begin{figure*}
    \centering
    \includegraphics[width=\linewidth]{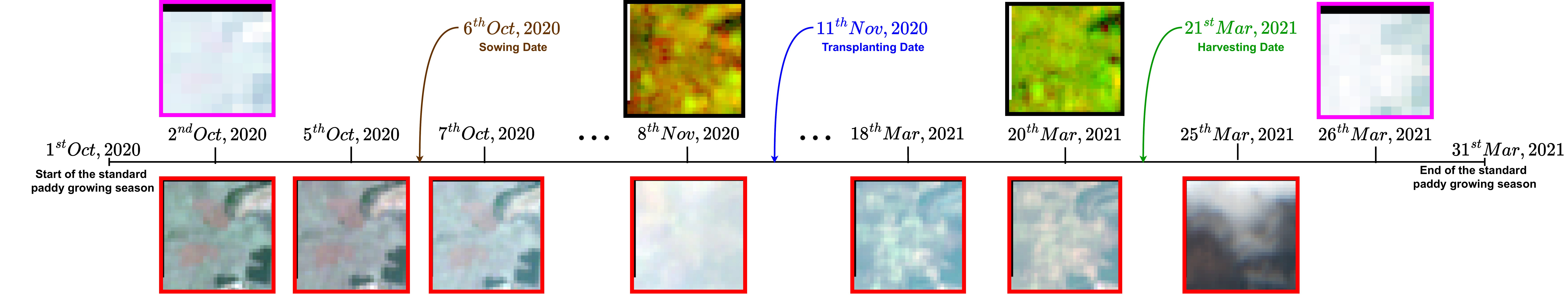}
    \caption{The axis denotes the individual days from start of the paddy growing season till the end of that season. We visualize a few image patches acquired on the corresponding dates from different satellites. Images with black boundaries denote images acquired from Landsat-8, pink boundaries denote Sentinel-1 and red boundaries denote Sentinel-2. Multiple observations from different satellites are observed on various dates as shown in the figure. In this case, there are no images on the sowing, transplanting and the harvesting dates.}
    \label{fig:sample-images}
\end{figure*}

Figure \ref{fig:dataset-analysis} presents the statistics of the complete dataset. It includes the various types of crops that are cultivated in the Cauvery Delta region in Tamil Nadu, the area of each plot in acres, the standard seasons (Table \ref{tab:standardseasons}) mapped using \cite{tnau} that are observed on the field as well, and the total number of samples in each district. Figure \ref{fig:sample-images} demonstrates a sample input from the dataset. Few images acquired on the dates mentioned on the axis are visualized in the figure along with the collected phenology dates. As in this example sample, there are cases where multiple images from different satellites are available on the same day but no images on the sowing, transplanting and harvesting are available. Tables \ref{tab:tasks} \& \ref{tab:characteristics} show a detailed comparison between the recent literature and our proposed dataset \dataset{}.

\begin{table}[h]
  \begin{center}
    {\small{
\begin{tabular}{lcc}
\toprule
\textbf{Season} & \textbf{Sowing Month} & \textbf{Duration (days)} \\
\midrule
Navarai & Dec. $-$ Jan. & 120 \\
Sornavari & Apr. $-$ May. & 120 \\
Early Kar & Apr. $-$ May. & 120 \\
Kar & May. $-$ June & 120 \\
Kuruvai & June $-$ July & 120 \\
Early Samba & July $-$ Aug. & 135 \\
Samba & Aug. & 180 \\
Late Samba & Sep. $-$ Oct. & 135 \\
Thaladi & Sep. $-$ Oct. & 135 \\
Late Pishanam & Sep. $-$ Oct. & 135 \\
Late Thaladi & Oct. $-$ Nov. & 120 \\
\bottomrule
\end{tabular}
}}
\end{center}
\caption{Standard growing seasons for paddy cultivation in the Cauvery Delta region obtained from Tamil Nadu Agricultural University (TNAU) \cite{tnau}. The time-series dataset is created using the maximum duration of the growing paddy season.}
\label{tab:standardseasons}
\end{table}

\begin{table*}
  \begin{center}
    {\small{
\begin{tabular}{lcccccccc}\toprule
\textbf{Tasks} & \textbf{\cite{sustainbench}} &\textbf{\cite{radiant}} &\textbf{\cite{agriculture-vision}} &\textbf{\cite{pixel-set}} &\textbf{\cite{pastis-r}} &\textbf{\cite{cropharvest}} &\dataset{} \\\midrule
Crop Type Semantic Segmentation & \cmark & \cmark &  & \cmark & \cmark & \cmark & \cmark \\
Crop Type Panoptic Segmentation &  &  &  &  & \cmark &  & \cmark \\
Cropland Segmentaion & \cmark & \cmark &  &  &  &  &  \\
Cropping Date Prediction &  &  &  &  &  &  & \cmark \\
Crop Yield Prediction & \cmark &  &  &  &  &  & \cmark \\
Field Delineation & \cmark &  & \cmark &  &  &  & \cmark \\
Crop Anomaly Detection &  &  & \cmark &  &  &  &  \\
\arrayrulecolor{black!30} \midrule 
Multi Image Super Resolution &  &  &  &  &  &  & \cmark \\
Cross-Satellite \& Cross-Sensor Fusion &  &  &  &  & \cmark &  & \cmark \\
Synthetic Band Generation &  &  &  &  & \cmark &  & \cmark \\
HR prediction using LR images &  &  &  &  &  &  & \cmark \\
\arrayrulecolor{black}\bottomrule
\end{tabular}
}}
\end{center}
\caption{A comparison of \dataset{} with related datasets based on the tasks that can be performed using them. The bottom $4$ tasks are not only related to the agricultural domain but are also applicable for remote sensing community.}
\label{tab:tasks}
\end{table*}

\begin{table*}
  \begin{center}
    {\small{
\begin{tabular}{lcccccccc}\toprule
\textbf{Characteristics} & \textbf{\cite{sustainbench}} &\textbf{\cite{radiant}} & \textbf{\cite{pastis-r}} & \textbf{\cite{agriculture-vision}} & \dataset{} \\
\midrule
Time Series data & \cmark & \cmark & \cmark &  & \cmark \\
Multiple annotations for the same plots &  &  &  & \cmark & \cmark \\
Annotations at multiple resolutions &  &  & &  & \cmark \\
Multi-Sensor data for all tasks & & & \cmark &  & \cmark \\
Consistent with the standard cropping practises &  &  &  &  & \cmark \\
Number of time-series samples & $[1,966-10,332]$ & Variable & $2,433$ & NA & $2,398$ \\
\arrayrulecolor{black}\bottomrule
\end{tabular}
}}
\end{center}
\caption{A comparison of \dataset{} with the related datasets (that can be used for multiple tasks) based on their characteristics.}
\label{tab:characteristics}
\end{table*}

\subsection*{Algorithm}
Algorithm \ref{alg:framework} executes some methods which are not defined in the main paper. We discuss the working of these methods in this section.
\begin{itemize}
    \item \textbf{\textit{tif\_to\_numpy\_sequence($P$)}:} This methods takes Path to the \textit{.tif} images $P$ as input and returns a time-series numpy array. These numpy arrays are generated using the proposed method in Section \ref{sec:methodology}.
    \item \textbf{\textit{get\_cropped\_inputs($\mathcal{X}$)}:} The time-series images $\mathcal{X}$ can be of any arbitrary dimension and the objective of this method is to create image patches of a fixed dimension that is to be inputted into the model. 
    \item \textbf{\textit{predict(task, $x$)}:} This method loads the pretrained model for the defined \textit{task} and generates predictions on the input data $x$.
    \item \textbf{\textit{merge\_cropped\_inputs($y$)}:} This method merges the prediction patches $y$ to form a single image of the original dimension of $\mathcal{X}$.
    \item \textbf{\textit{get\_paddy\_pixel\_indices($crop\_cover$)}:} This method returns the indices of the pixels where the pixels are predicted to be paddy, i.e. $pixel\_value \geq 0.5$
    \item \textbf{\textit{get\_patch(mid=(row, col), dim=(h, w), $\mathcal{X}$)}:} This method creates a small patch of dimension $h \times w$ from $\mathcal{X}$ centered at $(row, col)$. 
    \item \textbf{\textit{get\_actual\_season(patch, sow\_dt, harvest\_dt)}:} This method maks out all the images from $patch$ that are not a part of the actual season length, i.e. all the images before the sowing date $sow\_dt$ and after the harvesting date $harvest_dt$ are zeroed out.
\end{itemize}

\section*{Experiments}
\subsection*{Evaluation Metric}
\textbf{Crop Type Segmentation:} We use a per-pixel accuracy for evaluating the semantic segmentation task for the crop type. All the unknown pixels are masked out before calculating the metric.
\begin{equation}
    Accuracy = \frac{\textit{Number of correctly classified pixels}}{\textit{Total number of pixels}}
\end{equation}

\textbf{Phenology Dates Prediction:} We use Root Mean Squared Error (RMSE) as the evaluation metric for this task. We compute this metric for each individual phenology date, i.e. sowing, transplanting and harvesting dates. The RMSE signifies the average deviation (in days) from the actual cropping dates. 

\begin{align}
    RMSE &= \sqrt{\frac{\sum_{i=1}^N(y^i_{target} - y^i_{pred})^2}{N}} \\
\end{align}

\textbf{Crop Yield Prediction:} We again use the RMSE metric for this task. We also compute the range of error \% for the predicted yield, which signifies the \% deviation (in kg/acre) from the average yield (Table \ref{tab:error-range}), i.e.

\begin{align}
    Error \% &= \frac{RMSE}{\mu} \\
    \textit{where } \mu &= \textit{Mean of the target labels} \nonumber
\end{align}

\begin{table}
 \centering
 \begin{tabular}{cc}
  \toprule
  \textbf{Yield Range (kg/acre)} & \textbf{Error Range (\%)} \\
  \midrule
  $400 $-$ 450$ & $21.66\% $-$ 24.37\%$ \\
  $450 $-$ 500$ & $24.37\% $-$ 27.07\%$ \\
  $500 $-$ 550$ & $27.07\% $-$ 29.78\%$ \\
  $550 $-$ 600$ & $29.78\% $-$ 32.49\%$ \\
  \bottomrule
 \end{tabular}
 \caption{Range of the error \%}
 \label{tab:error-range}
\end{table}

\subsection*{Experimental Setup}
We create the training set and the test set using an 80/20 split strategy. Further 20\% of the training samples are held out for the validation set. 
All the tasks were performed using the same training configurations and no hyperparameter tuning is done as of now. We train the models for $130$ epochs with the learning rate $0.003$ and weight decay $0.01$. We use a $50\%$ dropout strategy as done by \cite{croptypemapping}.

\subsection*{Results}
We experimented with different band combinations from a single satellite and cross-satellite fusion to study the effects of spatial and spectral resolution. Even though we did not conduct an exhaustive experiment to find the best band combinations, the number of experiments we conducted was very high. Therefore, we present detailed experimentation on the Crop Yield Prediction task only so that the applicability of the actual growing season can also be demonstrated. We report only the best results obtained on Crop Type Segmentation and Phenology Date Prediction in Table \ref{tab:results}.
We compute the Normalized Difference Vegetation Index (NDVI), Green Chlorophyll Vegetation Index (GCVI), Red-Edge Chlorophyll Vegetation Index (RECI) and Ratio Vegetation Index (RVI) for each pixel in the input image and average them over pixels to get vegetation index for each timestep. We train a simple linear regression model for Phenology Dates \& Yield Prediction and consider the best-obtained results as the baseline for further experimentation (Table \ref{tab:yield-baseline}). We use the abbreviations L8, S1 and S2 to denote Landsat-8, Sentinel-1 and Sentinel-2, respectively. To denote the bands, we use the initials of each band. For example, the ultra-blue band is denoted by UB, red-edge 1 by RE1, and so on.

\begin{table*}
    \centering
    \begin{tabular}{c|ccc|c}
        \toprule
         \textbf{Task} & \textbf{L8} & \textbf{S1} & \textbf{S2} & \textbf{Metric}  \\
         \midrule
        Crop Type Mapping & $-$ & $-$ & All & 88.45\% \\
        Sowing Date Prediction& Ultra-Blue & $-$ & All & $\pm$ 2.578 days \\
        Transplanting Date Prediction & All Spectral & $-$ & $-$ & $\pm$ 3.631 days \\
        Harvesting Date Prediction & $-$ & VV & $-$ & $\pm$ 6.421 days \\
        Yield Prediction & Ultra-Blue & $-$ & All & $\pm$ 412.111 kg/acre  \\
        \bottomrule
    \end{tabular}
    \caption{Best results obtained for each task.}
    \label{tab:results}
\end{table*}


\begin{table*}
 \centering
 \begin{tabular}{c|ccc|ccc}
  \toprule
  \multicolumn{1}{l}{}  & \multicolumn{3}{c}{\textbf{Standard Growing Season}}   & \multicolumn{3}{c}{\textbf{Actual Growing Season}}  \\
  \midrule
    & \textbf{L8} & \textbf{S2} & \textbf{S1}    & \textbf{L8}    & \textbf{S2}    & \textbf{S1}    \\
  \textbf{NDVI}  & \cellcolor{yellow!40} 561.068  & \cellcolor{yellow!40} 524.937 & $-$  & 561.323    & \cellcolor{yellow!40} 508.372 & $-$  \\
  \textbf{GCVI}  & 568.240    & 547.820    & $-$  & 563.245 & 535.319    & $-$  \\
  \textbf{RECI}  & 566.880    & 542.773    & $-$ & \cellcolor{yellow!40} 541.162 & 539.146    & $-$  \\
  \textbf{RVI}   & $-$  & $-$  & \cellcolor{yellow!40} 584.727 & $-$  & $-$  & \cellcolor{yellow!40} 584.862 \\
  \bottomrule
 \end{tabular}
 \caption{Test RMSE values for Crop Yield Prediction using different vegetation indices. For each timestep, the vegetation index was averaged over all the pixels. The sequence of these vegetation indices was fed into a simple Linear Regressor to predict the crop yield. \colorbox{yellow!40} {Highlighted} rows represent the baseline results for each satellite.}
 \label{tab:yield-baseline}
\end{table*}

\begin{table*}
 \centering
 \begin{tabular}{c|cc|cc}
  \toprule
  \multicolumn{1}{l}{} & \multicolumn{2}{c}{\textbf{Standard Growing Season}} & \multicolumn{2}{c}{\textbf{Actual Growing Season}} \\
  \midrule
   & \textbf{L8} & \textbf{S2}  & \textbf{L8} & \textbf{S2} \\
  \textbf{R,G,B} & \cellcolor{red!20} 489.057  & \cellcolor{red!20} 490.022 & \cellcolor{red!20} 456.429 & \cellcolor{red!20} 466.212 \\
  \textbf{R,G,B,NIR} & 518.745 & 465.23 & \cellcolor{red!20} 453.882 & \cellcolor{red!20} 460.483 \\
  \textbf{R,NIR} & 542.578 & 480.168 & 482.147 & 440.691 \\
  \textbf{R,B,NIR} & 517.07 & 451.825 & 470.038 & 438.742 \\
  \textbf{G,NIR} & 525.819 & \cellcolor{red!40} 459.484 & 471.74 & \cellcolor{red!40} 464.292 \\
  \textbf{R,G,NIR} & 523.308 & 468.197 & 466.192 & 459.982 \\
  \textbf{SWIR1} & 532.064 & 448.105 & 478.838 & 442.448 \\
  \textbf{SWIR2} & 526.416 & 452.454 & 476.187 & 433.210 \\
  \bottomrule
 \end{tabular}
 \caption{Test RMSE values for Crop Yield Prediction using the common spectral bands available in all the different satellites. It is to be noted that all the experiments for yield prediction using a DL-based technique contain NDVI and GCVI as additional channels. \colorbox{red!40} {Highlighted} rows represent the experiment where the employment of actual growing season degraded the model performance. \colorbox{red!20}{Highlighted} rows represent the experiments where higher-spatial resolution images performed worse than the lower spatial resolution image with same spectral resolution. Both  of these scenarios are the failure cases of our initial assumptions.}
 \label{tab:yield-common}
\end{table*}

\begin{table*}
    \centering
    \begin{tabular}{cccc}
    \toprule
     & \textbf{Band combination} & \textbf{Standard Growing Season} & \textbf{Actual Growing Season} \\
     \midrule
    \textbf{L8} & \textbf{All Spectral} & 543.878 & 498.671 \\
    \textbf{L8} & \textbf{All}  & 532.994  & 458.766 \\
    \textbf{L8} & \textbf{UB} & 526.876  & 459.785 \\
    \textbf{L8} & \textbf{Thermal}      & 530.499 & 501.563 \\
    \textbf{L8} & \textbf{UB,Thermal}   & 537.722 & 497.534 \\
    \midrule
    \textbf{S2} & \textbf{All}  & 429.856 & \cellcolor{green!40} 414.228  \\
    \textbf{S2} & \textbf{RE1}  & \cellcolor{red!40} 424.670 & \cellcolor{red!40} 437.573  \\
    \textbf{S2} & \textbf{RE2}  & 479.990 & 436.3   \\
    \textbf{S2} & \textbf{RE3}  & \cellcolor{red!40} 441.378 & \cellcolor{red!40} 478.436  \\
    \textbf{S2} & \textbf{RE4}  & \cellcolor{red!40} 428.014 & \cellcolor{red!40} 465.341  \\
    \textbf{S2} & \textbf{RE(1-4)} & 461.786  & 448.379 \\
    \midrule
    \textbf{S1} & \textbf{All}  & 531.448 & 501.715 \\
    \textbf{S1} & \textbf{VV/VH}& 550.169 & 531.579 \\
    \textbf{S1} & \textbf{VV}& 553.304  & 535.112 \\
    \textbf{S1} & \textbf{VH}& 539.328  & 504.163 \\
    \textbf{S1} & \textbf{VV, Angle}    & 540.537 & 483.269 \\
    \textbf{S1} & \textbf{VH, Angle}    & 545.174 & 485.539 \\
    \bottomrule
    \end{tabular}
 \caption{Test RMSE values for Crop Yield Prediction using the different spectral bands from various satellites. Again, we add NDVI and GCVI as additional channels for these experiments. \colorbox{red!40} {Representation} is the same as explained in Table \ref{tab:yield-uncommon}. \colorbox{green!30}{Highlighted} cell represents the best performance of the model using only a single satellite.}
 \label{tab:yield-uncommon}
\end{table*}

\begin{table*}
 \centering
 \begin{tabular}{ccc|c} 
    \toprule
     \textbf{L8} & \textbf{S1} & \textbf{S2} & \textbf{Test RMSE} \\
     \midrule
     All & All & All & \cellcolor{yellow!40} 440.362 \\
     UB, Thermal & VV, Angle & All  & \color{red} 466.793 \\
     UB, Thermal & VH, Angle & All& 437.749 \\
     UB, Thermal    & VH & All& 416.777 \\
     UB & VV & RED, RE2, NIR & \color{red} 482.664 \\
     UB & VV & RED, BLUE, RE2, NIR & \color{red} 441.247 \\
     \rowcolor{green!10} UB & VV & RE2  & 415.643 \\
     UB & VH & RE2 & 429.371 \\
     UB & VH & SWIR2 & \color{red} 454.282 \\
     UB & VH & SWIR1, SWIR2 & \color{red} 465.055 \\
     UB & VH & RE1, RE2   & \color{red} 442.811 \\
     UB, Thermal & $-$ & All  & 413.049 \\
     Thermal & $-$  & All & 414.513 \\
     UB & $-$ & All  & \cellcolor{green!40} 412.111 \\
     \bottomrule
 \end{tabular}
\caption{Test RMSE values for Crop Yield Prediction using the multi-sensor data from multiple satellites. Again, we add NDVI and GCVI as additional channels for these experiments. Similar to Table \ref{tab:yield-baseline}, \colorbox{yellow!40} {row} represents the baseline result for cross-satellite fusion experiments. \colorbox{green!40}{Highlighted} cell represents the best performance of the model. Even though the performance of \colorbox{green!10}{row} is slightly lower than that of the best model, it is worth noticing that this result is obtained using only a single sensor from each satellite. For deployments with less computational power, such a band combination is more sensible. Cells with \textcolor{red}{red} text color represent the experiments that were not performing better than the baseline method for fusion experiments.}
 \label{tab:yield-fusion}
\end{table*}